\documentclass[sigconf,screen]{acmart}

\setcopyright{acmlicensed}
\acmDOI{10.1145/3650212.3680380}
\acmYear{2024}
\copyrightyear{2024}
\acmISBN{979-8-4007-0612-7/24/09}
\acmConference[ISSTA '24]{Proceedings of the 33rd ACM SIGSOFT International Symposium on Software Testing and Analysis}{September 16--20, 2024}{Vienna, Austria}
\acmBooktitle{Proceedings of the 33rd ACM SIGSOFT International Symposium on Software Testing and Analysis (ISSTA '24), September 16--20, 2024, Vienna, Austria}
\acmSubmissionID{issta24main-p1278-p}
\received{2024-04-12}
\received[accepted]{2024-07-03}

\usepackage{amsthm}
\usepackage{bbm}
\usepackage{theoremref}
\usepackage{algpseudocode}
\usepackage{tikz}

\newcommand{\neuronrepair}[0]{\textsc{NeuFair}\xspace}

\usepackage{hyperref}
\usepackage{subcaption}
\newsavebox{\mybox}

\usepackage{amsthm}
\usepackage{mdframed}
\usepackage{centernot}
\usepackage{comment}
\usepackage{microtype}
\usepackage{siunitx}
\usepackage{dsfont}

\newcommand{\Dd}{\mathcal{D}}

\newcommand\vd[2]{d_{i, p}}
\newcommand{\set}[1]{\left\{ #1 \right\}}

\newcommand{\R}{\mathbb R}

\newcommand{\toolname}{\textsc{NeuFair}\xspace}
\usepackage{tikz}

\newtheorem{definition}{Definition}[section]

\definecolor{gold}{rgb}{0.99,0.78,0.07}
\usetikzlibrary{arrows,shapes,snakes,automata,backgrounds,positioning,decorations.pathmorphing,
	decorations.markings,calc}

\tikzstyle{dtreenode}=[draw=blue!10!gray,rounded rectangle, minimum size=5mm,fill=blue!10!white]
\tikzstyle{dtreeleaf}=[draw=black!60,minimum width=1cm,minimum height=0.4cm,rectangle,fill=blue!50!white]
\tikzset{every loop/.style={looseness=7}}
\tikzset{
	gluon/.style={decorate,draw=black,
		decoration={coil,amplitude=1pt, segment length=5pt}}
}
\tikzset{
	gluon1/.style={decorate,draw=black,
		decoration={coil,amplitude=3pt, segment length=3pt}}
}
\tikzset{
	gluonew/.style={decorate,draw=black,
		decoration={coil,amplitude=1pt, segment length=2pt}}
}

\tikzset{bicolor/.style args={#1 and #2 and #3}{
		path picture={
			\tikzset{rounded corners=0}
			\fill [#1] (path picture bounding box.south west)
			rectangle
			($(path picture  bounding box.north west)!#3!(path picture bounding
			box.north east)$);
			\fill [#2]
			($(path picture bounding box.south west)!#3!(path picture bounding
			box.south east)$)
			rectangle (path picture bounding box.north east);
}}}

\tikzset{tricolor/.style args={#1 and #2 and #3 and #4 and #5}{
		path picture={
			\tikzset{rounded corners=0}
			\fill [#1] (path picture bounding box.south west)
			rectangle
			($(path picture  bounding box.north west)!#4!(path picture bounding
			box.north east)$);
			\fill [#2]
			($(path picture bounding box.south west)!#4!(path picture bounding
			box.south east)$)
			rectangle
			($(path picture  bounding box.north west)!#5!(path picture bounding
			box.north east)$);
			\fill [#3]
			($(path picture bounding box.south west)!#5!(path picture bounding
			box.south east)$)
			rectangle (path picture bounding box.north east);
}}}

\usepackage[many]{tcolorbox}
\tcbuselibrary{listings,skins}

\lstdefinestyle{mystyle}{
  xleftmargin=0pt,
   basicstyle={\footnotesize\ttfamily},
   aboveskip=3mm,
   belowskip=3mm,
   keywordstyle=\bfseries,
   showstringspaces=false,
  escapechar=?,
  language=Java
}
\definecolor{code_indent}{HTML}{CCCCCC}

\newtcblisting{mylisting}[2][]{
  arc=0pt, outer arc=0pt,
  listing only,
  listing style=mystyle,
  title={\large #2},
  #1
}

 \definecolor{dkgreen}{rgb}{0,0.6,0}
 \definecolor{gray}{rgb}{0.5,0.5,0.5}
 \definecolor{mauve}{rgb}{0.58,0,0.82}

\definecolor{cadmiumgreen}{rgb}{0.0, 0.42, 0.24}
\definecolor{verde}{rgb}{0.25,0.5,0.35}
\definecolor{jpurple}{rgb}{0.5,0,0.35}
\definecolor{darkgreen}{rgb}{0.0, 0.2, 0.13}

\usepackage[ruled,vlined,linesnumbered,lined]{algorithm2e}
\usepackage{algpseudocode}

\usepackage{mathtools,xparse}

\usepackage{tabu}
\usepackage{multirow}

\usepackage{pifont}
\usepackage{enumerate}
\usepackage[shortlabels]{enumitem}

\usepackage{pgfplotstable}
\usepackage{pgfplots}

\usepackage[noframe]{showframe}
\usepackage{framed}

 {\endMakeFramed}
 \definecolor{shadecolor}{gray}{0.85}

\definecolor{bgblue}{RGB}{245,243,253}
\definecolor{ttblue}{RGB}{91,194,224}

\mdfdefinestyle{mystyle}{%
  rightline=true,
  innerleftmargin=10,
  innerrightmargin=10,
  outerlinewidth=3pt,
  topline=false,
  rightline=false,
  bottomline=false,
  skipabove=\topsep,
  skipbelow=false
}

\newtcolorbox{myboxi}[1][]{
  breakable,
  title=#1,
  colback=white,
  colbacktitle=white,
  coltitle=black,
  fonttitle=\bfseries,
  bottomrule=0pt,
  toprule=0pt,
  leftrule=3pt,
  rightrule=3pt,
  titlerule=0pt,
  arc=0pt,
  outer arc=0pt,
  colframe=black!50,
}

\newtcolorbox{myboxii}[1][style=mystyle]{
  breakable,
  freelance,
  colback=white,
  colbacktitle=white,
  coltitle=black,
  fonttitle=\bfseries,
  bottomrule=0pt,
  boxrule=0pt,
  colframe=white,
  after skip=0pt,
  overlay unbroken and first={
    \draw[white!75!black,line width=3pt]
    ([yshift=-9pt]frame.north west) --
    ([yshift=9pt]frame.south west);
  },
  }

\author{Vishnu Asutosh Dasu}
\orcid{0000-0002-1849-1288}
\affiliation{%
  \institution{Pennsylvania State University}
  \city{State College}
  \country{USA}
}
\email{vdasu@psu.edu}

\author{Ashish Kumar}
\orcid{0000-0001-8773-2084}
\affiliation{%
  \institution{Pennsylvania State University}
  \city{State College}
  \country{USA}
}
\email{azk640@psu.edu}

\author{Saeid Tizpaz-Niari}
\orcid{0000-0002-1375-3154}
\affiliation{%
  \institution{University of Texas at El Paso}
  \city{El Paso}
  \country{USA}
}
\email{saeid@utep.edu}

\author{Gang Tan}
\orcid{0000-0001-6109-6091}
\affiliation{%
  \institution{Pennsylvania State University}
  \city{State College}
  \country{USA}
}
\email{gtan@psu.edu}

\ccsdesc[500]{Software and its engineering~Search-based software engineering}
\ccsdesc[500]{Computing methodologies~Machine learning}

\keywords{Machine Learning, Bias Mitigation, AI Ethics}

\begin{document}

\title{\neuronrepair: Neural Network Fairness Repair with Dropout}

\begin{abstract}
  This paper investigates neuron dropout as a post-processing bias mitigation method for deep neural networks (DNNs). Neural-driven software solutions are increasingly applied in socially critical domains with significant fairness implications. While DNNs are exceptional at learning statistical patterns from data, they may encode and amplify historical biases. Existing bias mitigation algorithms often require modifying the input dataset or the learning algorithms.
  We posit that prevalent dropout methods may be an effective and less intrusive approach to improve fairness of pre-trained DNNs during inference. However, finding the ideal set of neurons to drop is a combinatorial problem.

  We propose \neuronrepair, a family of post-processing randomized algorithms
  that mitigate unfairness in pre-trained DNNs via dropouts during inference. Our randomized search is guided by an objective to minimize discrimination while maintaining the model's utility. We show that \neuronrepair~is efficient and effective in improving fairness (up to 69\%) with minimal or no model performance degradation. We provide intuitive explanations of these phenomena and carefully examine the influence of various hyperparameters of \neuronrepair~on the results. Finally, we empirically and conceptually compare \neuronrepair to different state-of-the-art bias mitigators.
\end{abstract}

\maketitle

\section{Introduction}
\label{sec:introduction}
Artificial intelligence (AI), increasingly deployed with deep neural network (DNN) components, has become an integral part of modern software solutions that assist in socio-economic and legal-critical decision-making processes such as releasing patients~\cite{healthcareapp},
identifying loan defaults~\cite{Home-Credit-Default-Risk}, and
detecting tax evasion~\cite{nyt2023taxaudit}. 

Despite many advances made possible by AI,
some challenges require understanding the dimensions and implications of deploying AI-driven
software solutions. One such concern about the trustworthiness of AI is discrimination. Unfortunately, there are plenty of fairness defects in real systems. Parole decision-making software was found to harm black and Hispanic defendants by falsely predicting a higher risk of recidivism than for non-Hispanic white defendants~\cite{compas-article}; Amazon's hiring algorithm disproportionately rejected more female applicants than male applicants~\cite{Amazon-same-day-delivery}; and data-driven auditing algorithm selected black taxpayers with earned income tax credit claims (EITC) at much higher rates than other racial groups for an audit~\cite{nyt2023taxaudit}. As evidenced by these examples, resulting software may particularly disadvantage minorities and protected groups and be found non-compliant with laws such as the US Civil Rights Act~\cite{blumrosen1978wage}. Hence, helping programmers and users to mitigate unfairness in social-critical data-driven software systems is crucial to ensure inclusion in our modern, increasingly digital society.

The software engineering (SE) community has spent significant efforts to address discrimination in the automated data-driven software solutions~\cite{FairnessTesting,
10.1145/3338906.3338937,udeshi2018automated,ADF,chakraborty2020fairway}. Fairness has been treated as a critical meta-property that requires analysis beyond functional correctness and measurements beyond prediction accuracy~\cite{10.1145/3236024.3264838}. Thus, the community presents various testing~\cite{FairnessTesting,10.1145/3338906.3338937,zhang2020white},
debugging~\cite{gohar2023understanding,10.1109/ICSE48619.2023.00136,yu2023fairlayml}, and mitigation~\cite{zhang2021ignorance,10.1145/3468264.3468536} techniques to address fairness defects in data-driven software. 

Broadly, fairness mitigation can be applied in the pre-processing (e.g., increasing the representations of an under-represented group by generating more data samples for them), in-processing (e.g., changing the loss function to include fairness constraints during training), or post-processing (e.g., changing the logic of a pre-trained model) stage. 
However, when the decision logic of AI is encoded via DNNs, it becomes challenging to mitigate unfairness due to its black-box uninterpretable nature. 

We posit that a subset of neurons in a neural network disparately contributes to unfairness. Removing these neurons during inference as a post-processing operation on a trained DNN can improve fairness.
Therefore, prevalent techniques such as dropout methods~\cite{10.5555/2627435.2670313,srivastava2013improving}---a process of randomly dropping neurons during training---might be effective in mitigating unfairness in \textit{pre-trained} DNN models. While the dropout strategy has been significantly used to prevent over-fitting during the training of DNNs, to the best of our knowledge, this is the first work to systematically leverage dropout methods as a post-processing bias mitigation method for improving the fairness of \textit{pre-trained DNN} models.

However, finding the optimal subset of dropout neurons is an intractable combinatorial problem that requires an exhaustive search over all possible combinations of neuron dropouts. To overcome the computational challenges, we explore
the class of randomized algorithms with Markov chain Monte Carlo (MCMC)
strategies to efficiently explore the search space with statistical guarantees.
In doing so, we pose the following research questions:

\vspace{0.25 em}
\noindent \emph{RQ1. How successful are randomized algorithms in repairing the unfairness of DNNs via dropouts?}

\vspace{0.25 em}
\noindent \emph{RQ2. Are there dropout strategies that improve model fairness and utility together?}

\vspace{0.25 em}
\noindent \emph{RQ3. What are the design considerations of search algorithms for efficient and effective dropouts?}

\vspace{0.25em}
\noindent \emph{RQ4. How do dropout strategies compare to the state-of-the-art post-processing (bias) mitigators?}
 
To answer these research questions, we present \toolname (Neural Network Fairness Repair): a set of randomized 
search algorithms to improve the fairness of DNNs via inference-time neuron dropouts. 
We design and implement simulated annealing (SA) and random walk (RW) strategies that
efficiently explore the state-space of neuron dropouts where we encode the frontiers of fairness and utility in a cost function. 


We evaluate \toolname over $7$ deep neural network benchmarks trained
over $5$ socially critical applications with significant fairness implications. We found
that \toolname can improve fairness (up to 69\%) with minimal utility degradation in most cases. We also report a pathological
case and reasons behind a failure of \toolname. We also observe that \toolname can simultaneously improve
fairness and utility and provide intuitive explanations of such phenomena. Furthermore,
we examine different hyperparameter configuration options of randomized algorithms.
While some hyperparameters always influence fairness with positive or negative impacts, we detect a hyperparameter that defines a trade-off between explorations
and exploitation that should be tuned as a constant variable in the cost function
for each benchmark. Finally, we show the effectiveness of the SA algorithm, compared to the RW and
the state-of-the-art post-processing (bias) mitigator~\cite{10.1109/ICSE48619.2023.00136}.

In summary, the key contributions of this paper are:
\begin{enumerate}
  \item \textbf{Inference-Time dropout for fairness}. To the best of our knowledge, we present the first dropout method of bias mitigation over pre-trained deep neural networks,
  \item \textbf{Randomized algorithms for fairness}.
  We create a well-defined framework to formulate the combinatorial inference-time dropout problem in DNNs using randomized algorithms,
  \item \textbf{Experimental evaluations}. We implement the randomized algorithms in \toolname and
  evaluate their effectiveness and efficiency vis-a-vis the state-of-the-art techniques.
\end{enumerate}

\section{Background}
\label{sec:background}
In this section, we provide a background on the various model utility and fairness metrics. 

\subsection{Notions of Model Utility}

Given a binary classifier $h$, a set of features $X$, and predictions $h(X) = \{TP, TN, FP, FN\}$, we can define the following notions of model utility. In $h(X)$, TP, TN, FP, and FN denote the set of True Positives, True Negatives, False Positives, and False Negatives.  We drop the cardinality operator $\lvert \cdot \rvert$ in the following definitions for brevity.

\begin{equation*}
    \text{Accuracy}_h = \frac{TP+TN}{TP+TN+FP+FN}
\end{equation*}



\begin{equation*}
    \text{F1}_h = \frac{2*Precision*Recall}{Precision+Recall} = \frac{2*TP}{2*TP+FP+FN}
\end{equation*}



Accuracy can be used to gauge the overall performance of a classifier. However, accuracy is a poor metric for imbalanced datasets commonly used in fairness evaluations as the number of negative samples far outweighs the positive samples. For example, predicting all samples as negative (0 true positives) in the \textit{Bank} \cite{Dua:2019-bank} dataset yields an accuracy of 88\% but the F1 score would be 0 or undefined. Since F1 is defined as the harmonic mean of precision and recall, rather than the arithmetic mean, it penalizes performance significantly if either precision or recall is low.

\subsection{Notions of Model Fairness}

Consider a machine learning classifier $h$, a set of features $X$, sensitive features $A \subset X$, and a set of labels $Y$.
We can then define the following notions of fairness.

\begin{definition}[Demographic Parity \cite{agarwal2018reductions}]
The classifier $h$ satisfies Demographic Parity under a distribution over $(X,A,Y)$ if its prediction $h(X)$ is statistically independent of the sensitive feature $A$ i.e. $\mathbb{P}[h(X) = \hat{y} | A = a] = \mathbb{P}[h(X) = \hat{y}]$ for all $a, y$. For binary classification with $\hat{y} = \{0, 1\}$, this is equivalent to $\mathbb{E}[h(X) | A = a] = \mathbb{E}[h(X)]$ for all $a$.
\end{definition}

\begin{definition}[Equalized Odds \cite{hardt2016equality, agarwal2018reductions}]
The classifier $h$ satisfies Equalized Odds under a distribution over $(X, A, Y)$ if its prediction $h(X)$ is conditionally independent of the sensitive feature $A$ given the label $Y$ i.e. $\mathbb{P}[h(X) = \hat{y} | A = a, Y = y] = \mathbb{P}[h(X) = \hat{y} | Y = y]$ for all $a$, $y$, and $\hat{y}$. For binary classification with $\hat{y} = \{0, 1\}$, this is equivalent to $\mathbb{E}[h(X) | A = a, Y = y] = \mathbb{E}[h(X) | Y = y]$ for all $a$, $y$. 
\end{definition}


Equal Opportunity is a relaxed variant of Equalized Odds with $Y = 1$ \cite{hardt2016equality}. Equalized Odds require the true positive and false positive rates to be equal across all sensitive groups. However, Equal Opportunity only requires the true positive rate to be equal across all sensitive groups. 

Demographic Parity is the weakest notion of fairness, and Equalized Odds is the strongest. In our work, we use Equalized Odds as the fairness criterion. The disparity or unfairness for Equalized Odds is the Equalized Odds Difference (EOD), defined as the maximum absolute difference between the true and false positive rates across the sensitive groups. Mathematically, this is represented as



{
\small
\begin{equation*}
    EOD = \max \left(
    \begin{aligned}
        \lvert \mathbb{P}[h(X) = 1 | A = 0, Y = 1] - \mathbb{P}[h(X) = 1 | A = 1, Y = 1] \rvert, \\
        \lvert \mathbb{P}[h(X) = 1 | A = 0, Y = 0] - \mathbb{P}[h(X) = 1 | A = 1, Y = 0] \rvert
    \end{aligned}
    \right)
\end{equation*}
}

with sensitive features $A = \{0,1\}$.

\section{Problem Statement}
\label{sec:problem}

We consider pre-trained deep neural network (DNN) classifiers with the set of input variables $X$ partitioned into a protected set of variables (such as race, sex, and age) and non-protected variables (such as profession, income, and education). 
We further assume the output is a binary classifier that gives either favorable or unfavorable outcomes. 

\subsection{Syntax and Semantics of DNN}
A deep neural network (DNN) encodes a function $\Dd: X \to [0,1]^2$ where $X$ consists of the set of protected
attributes $X_1 \times X_2 \cdots \times X_m$ and non-protected attributes $X_{m+1} \cdots \times Y_{m+r}$.
The DNN model is parameterized by the input dimension $m{+}r$, the output dimension $2$, the depth of hidden layers $n+1$, and the weights of its hidden layers $W_0, W_1, \ldots, W_n$. We describe the hidden layers with $\mathcal{M} \gets [L_I, L_0, \ldots, L_n, L_O]$, where $L_I$ and $L_O$ are the input and output layers, respectively, and ${L_i}, \forall i \in [0,n]$, are the hidden layers. We assume that there exists a subset of neurons $\mathcal{N} \in {L_i}, \forall i \in [0,n]$ in the hidden layers that disparately contribute to unfairness.

Let $L_{i}$ be the output of layer $i$ that implements an affine mapping from the output of previous layer $L_{i-1}$ and its weights $W_{i-1}$ for $1 \leq i \leq n$ followed by a fixed non-linear activation unit (e.g., ReLU defined as $L_{i-1} \mapsto \max \set{W_{i-1}.L_{i-1}, 0}$) for $1 \leq i \leq n$. Let $L_i^{j}$ be the output of neuron $j$ at layer $i$ that
is $L_i^j(x) = \texttt{ReLU}\left(\sum_{j = 1} ^ {|L_i|} w_{i} L_{i-1}^j\right)$. 
The output is the likelihood of favorable and unfavorable outcomes. The predicted label is the index of the maximum likelihood,
$\Dd(x) = \max_{i} L_O(x)(i)$. 







\subsection{Inference Time Dropout for Fairness}
Dropout \cite{10.5555/2627435.2670313} is a technique proposed to improve the performance of DNNs by preventing overfitting. 
Dropout sets all $w_i$ to 0 for a random set of neurons in the hidden layers with some probability during training. Once training is complete, dropout is not used, and all the neurons in the DNN are utilized to make predictions. While dropout has been traditionally used to prevent overfitting during training,
we hypothesize that dropping neurons of the DNN during inference after training can significantly improve fairness with a minimal impact on performance. However, unlike traditional dropout, where a set of neurons are randomly dropped during training, we aim to identify a subset of neurons at the Pareto optimal curve of fairness-performance during inference.

We consider a binary vector as the neuron state with $s = \{0,1\}^{N}$,
where $N = \sum_{i=0}^{n} |L_i|$ and $s_i$ indicates whether
the neuron $i$ is dropped or not and $n$ is the number of layers. A pre-trained DNN model $\Dd$ does not
include any dropouts; hence, all the indicators are $0$. 



\begin{definition}[Desirable Dropout of Fairness vs. Utility]\label{def:problem}
Given a DNN model $\Dd$ trained over a dataset $X$;
the search problem is to infer a repaired DNN model $\Dd'$ by dropping
a subset of neurons $I$ 
in the binary neuron state, i.e., $s_i = 1$ for any $i \in I$, such that 
(1) the model bias (e.g., EOD) is maximally reduced, (2) the model performance
(e.g., F1-score) is minimally degraded, and (3) the model structure in 
terms of the numbers of inputs, outputs, and hidden layers remains the same as compared
to the original pre-trained model $\Dd$.
\end{definition}


We define desirable states as those states that have a good fairness-performance tradeoff. A brute-force search to find desirable neurons is exponential in the size of DNN as we have $2^{|N|}$ possible subsets.
The running time becomes prohibitively expensive, even for small DNNs.  To find the desirable subset of neurons, we explore different types of randomized algorithms to improve fairness via model inference dropout. 


\section{Approach}
\label{sec:approach}
Our approach comprises two randomized search algorithms, namely Simulated Annealing (SA) and Random Walk (RW).

\begin{figure*}
    \includegraphics[width=0.9\textwidth]{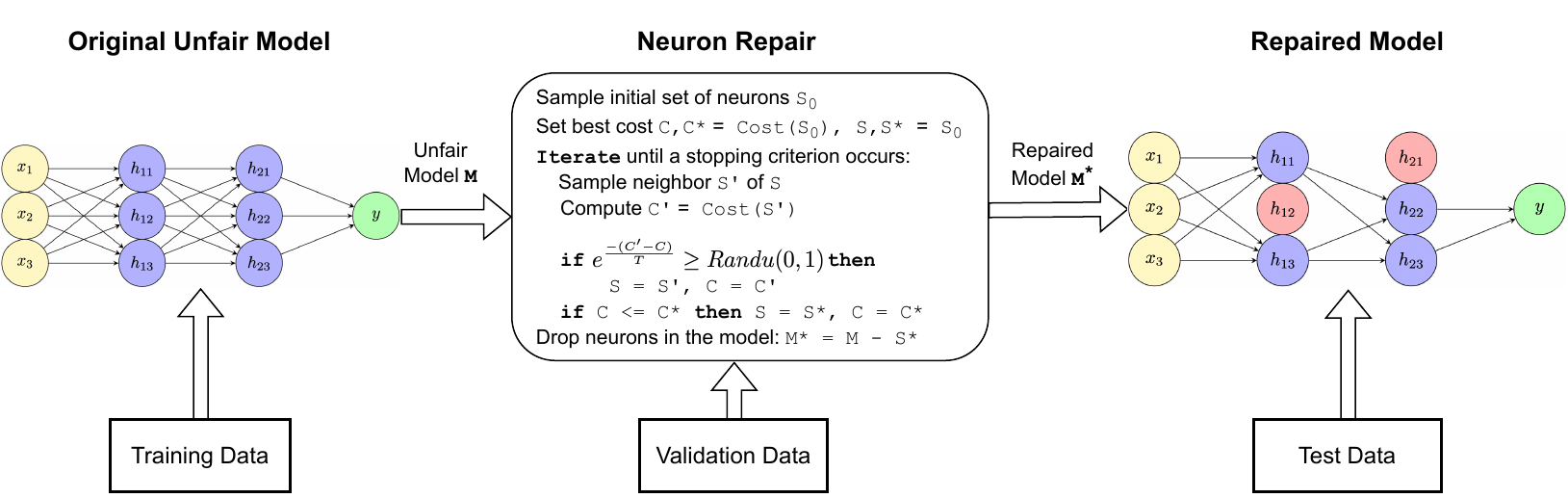}
    \caption{Overview of the \neuronrepair: a post-processing bias mitigation approach. }
    \label{fig:overview-approach}
\end{figure*}

\subsection{Simulated Annealing Search}

We formulate the problem of finding the desirable subset of neurons as a discrete optimization problem and solve it using Simulated Annealing. Simulated Annealing (SA) \cite{BertTsit93} is a probabilistic algorithm that finds the global minima (w.r.t some cost function) in a large search space with high probability. Algorithm~\ref{alg:gen_sa} presents a generic template to apply SA to a search space optimization problem. Figure~\ref{fig:overview-approach} overviews the steps in our bias mitigation
approach with the SA search. We now define the core concepts used in our SA algorithm.

\begin{algorithm}
\SetAlgoNlRelativeSize{0}
\caption{Generic Simulated Annealing (SA) Procedure}
\label{alg:gen_sa}

\KwIn{Search space $\mathcal{S}$, neighborhood relation $\Gamma$, cost function $c$, initial temperature $T_0$, initial state $s_0$}
\KwOut{Best solution found.}
\BlankLine

$T, s, s^*, m \gets T_0, s_0, s_0, 0$ \\

\While{`stopping criterion' is not met}{
    $s' \gets \text{Generate}(s)$
    
    Compute $\Delta E = \text{cost}(s') - \text{cost}(s)$
    
    \If{$\Delta E \le 0$}{
        Accept transition i.e. $s \gets s'$ \\
        Update best solution found i.e. $s^* \gets s'$
    }
    \Else{
        \If{$e^{-\Delta E / T_m} \ge \text{UniformSample}(0, 1)$}{
            Accept transition i.e. $s \gets s'$
        }
    }
    
    $T_{m+1} \gets \text{Update}(T_m)$ \\
    $m \gets m + 1$
}

\Return{$s^*$}
\end{algorithm}

\paragraph{Search Space}
For a DNN $\mathcal{M} \gets [L_I, L_0, \ldots, L_n, L_O]$ we define a state of our search space $\mathcal{S}$ as a binary sequence $s \in \{0, 1\} ^ N$, where $N = \sum_{i=0}^{n} |L_i|$\footnote{For memory reasons, we store a state as a decimal number rather than a $N$-bit binary sequence, where the binary sequence $s$ is mapped to the unique decimal number which has binary expansion $1s$. }. The $i$th element of $s$, denoted $s_i$ is $1$ if neuron $i$ is dropped and $0$ otherwise\footnote{The neurons are numbered according to some total order; the choice of total order is insignificant as long as we fix the mapping of a neuron to its position in the binary sequence.}.
For example, consider a DNN $\mathcal{M} \gets [L_I, L_0, L_1, L_O]$, where $|L_0| = 3$ and $|L_1| = 3$. Then, the state $s$ given by:
\begin{equation*}
    s = \underbrace{(0, 1, 0}_{L_0}, \underbrace{0, 0, 1)}_{L_1}
\end{equation*}
drops the second neuron in the first layer and the third neuron in the second layer. 

Instead of allowing our search space $\mathcal{S}$ to drop every possible subset of neurons possible (which would ensure $\mathcal{S}$ has size $2^N$), we restrict the size of $\mathcal{S}$ by fixing an upper and lower bound on the number of neurons that can be dropped from the DNN. Let $n_l$ and $n_u$ ($n_l \le n_u$) denote the minimum and maximum number of neurons allowed by our DNN, respectively. We can then formally define $\mathcal{S}$ as:
\begin{equation}
    \mathcal{S} := \{s \in \{0,1\}^N \; | \; n_l \le HW(s) \le n_u \}
\end{equation}
where $HW(s)$ denotes the hamming weight of $s$. Restricting our search space with a conservative estimate of the lower and upper bound would have a minimal impact on our ability to find the desirable subset, as dropping too many neurons will reduce the model utility to less than acceptable levels, whereas dropping too few neurons will improve fairness by only a marginal amount. 
However, asymptotically, a conservative estimate of the lower and upper bound still generates prohibitively large search spaces. With the bounds $n_l$ and $n_u$, the cardinality of the search space $|\mathcal{S}| = \sum_{i = n_l}^{n_u} {N \choose i} = \Omega(\frac{2^N}{\sqrt{N}})$ if $n_l < \frac{N}{2} < n_u$, which rules out brute-force as a viable option.



\paragraph{Neighborhood Relation and `Generate' subroutine.}
The neighborhood of any state $s \in \mathcal{S}$, denoted $\Gamma(s)$, is defined as the set of all states that are at a hamming distance of 1. Mathematically, this is defined as
\begin{equation*}
    \Gamma(s) := \{s' \in \mathcal{S} \; | \; HD(s,s') = 1 \}
\end{equation*}
where $HD(s,s')$ denotes the hamming distance between $s$ and $s'$. With our definition of the search space and neighborhood of a state, the entire search space graph can be viewed as a subset of the $N$-dimensional hypercube, where the vertices of the hypercube represent the states of our search space, and the edges of the hypercube represent the neighborhood relation. The \textit{Generate} subroutine on input $s \in \mathcal{S}$ uniformly samples a neighbor $s'$ from $\Gamma (s)$ and returns $s'$. This is equivalent to uniformly sampling an index position $i$ from $[1,n]$ and subsequently uniformly flipping bit $s_i$ in the binary sequence $s$ to get $s'$. We make the following observation on our underlying search space graph\footnote{The search space graph is a graph with $\mathcal{S}$ as the set of vertices and edges defined according to the neighborhood relation $\Gamma$ i.e. $(s,s')$ is an edge iff $s' \in \Gamma(s)$. As our neighborhood relation is symmetric, our search space graph is undirected.}:  
\begin{lemma}
    If $n_l < n_u$, then our search space graph is connected and has a diameter less than $N$. Moreover, the distance between any two states $s,s' \in \mathcal{S}$ is given by $HD(s,s')$.
\end{lemma}

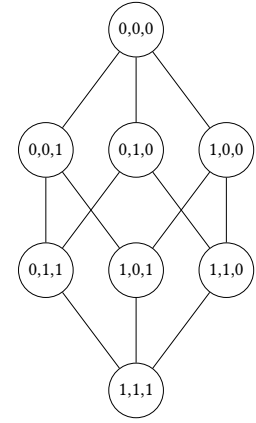
\begin{figure}
\centering
\scalebox{0.8}{\begin{tikzpicture}
    \pgfmathdeclarefunction{hammingweight}{3}{
        \pgfmathtruncatemacro{\hw}{abs(#1)+abs(#2)+abs(#3)}
        \pgfmathtruncatemacro{\result}{\hw}
        \pgfmathresult=\result
    }

    \node[draw, circle] (000) at (3, 0) {0,0,0};
    \node[draw, circle] (111) at (3, -6) {1,1,1};

    \node[draw, circle] (001) at (1.5, -2) {0,0,1};
    \node[draw, circle] (010) at (3, -2) {0,1,0};
    \node[draw, circle] (100) at (4.5, -2) {1,0,0};

    \node[draw, circle] (011) at (1.5, -4) {0,1,1};
    \node[draw, circle] (101) at (3, -4) {1,0,1};
    \node[draw, circle] (110) at (4.5, -4) {1,1,0};

    \draw (000) -- (001);
    \draw (000) -- (010);
    \draw (000) -- (100);
    \draw (001) -- (011);
    \draw (001) -- (101);
    \draw (010) -- (011);
    \draw (010) -- (110);
    \draw (100) -- (101);
    \draw (100) -- (110);
    \draw (011) -- (111);
    \draw (101) -- (111);
    \draw (110) -- (111);
\end{tikzpicture}}
\caption{3-dimensional Boolean hypercube that demonstrates the exponential search space of a neural network with 1 hidden layer having 3 neurons.}
\label{fig:example_space}
\end{figure}

For example, Figure~\ref{fig:example_space} shows the search space graph for a neural network with 1 hidden layer with 3 neurons, along with the neighbors and state transitions.

\paragraph{Cost Function}
 Our primary goal is to find a state $s \in \mathcal{S}$ which minimizes its unfairness score $EOD_s$. However, we do not wish to consider those states with a significant loss in model performance (measured using the F1 score of the model) compared to the original model; thus, we try to find an acceptable balance between the improvement in fairness and the loss in model performance. To realize this balance, we penalize our cost function with an additional term to artificially increase the cost if the state has a lower F1 score. We now formally define 
 our cost function, $\text{cost}(\cdot)$ as follows:
\begin{equation}
    \text{cost}(s) := EOD_s + p \cdot EOD_{s_0} \cdot \mathds{1} (F1_{s} < t F1_{s_0})
    \label{eq:cost_function}
\end{equation}
In the above equation, $s_0$ is the original state (the DNN with no dropout), $p \in \R_{\ge 0}$ is called the penalty multiplier, $t \in (0,1)$ is called the threshold multiplier, $EOD_s$ and $EOD_{s_0}$ are the unfairness scores of states $s$ and $s_0$ respectively, $F1_s$ and $F1_{s_0}$ are the F1 scores of the $s$ and $s_0$ respectively, and $\mathds{1}(\cdot)$ denotes the indicator function. The threshold multiplier $t$ determines the percentage loss in F1 score we will tolerate to improve fairness. The penalty multiplier $p$ discourages states with an F1 score less than the threshold by penalizing them with a multiple of the unfairness of the initial state. Our cost function formulation allows us to find the state with the minimum unfairness while maintaining a significantly higher F1.

\paragraph{Initial Temperature, Cooling Schedule, and `Update' subroutine}

The temperature $T$ of the SA procedure determines the probability with which we accept a positive transition (a transition where the cost is increased); the higher the temperature, the more likely the algorithm accepts such a transition. The cooling schedule refers to the function used to update the temperature after each iteration. We adopt the logarithmic cooling schedule \cite{hajek1988cooling} which has proven convergence guarantees \cite{mitra1986convergence}. According to the logarithmic cooling schedule, the temperature $T(\cdot)$ at iteration $m$ is defined as
\begin{equation}
    T_m = \frac{T_0}{\log(2 + m)}, \forall m \in \mathbb{Z}_{\ge 0}
\end{equation}
The \textit{Update} subroutine updates the temperature using the above formula. Using the convergence results for SA with a logarithmic cooling schedule \cite{mitra1986convergence}, we get the following result for our SA runs:
\begin{lemma}
    Let $p > 1$. If we set $T_0 \ge (1 + p \cdot EOD_{s_0})(n_u - n_l)$, then the probability that SA finds a global minima within $k > \lceil \frac{n_u - n_l}{2} \rceil $ iterations is greater than $1 - \frac{A}{(\frac{k}{n_u - n_l})^{c}}$ where $$c := \min{( \frac{1}{(n_u-n_l)N^{\lceil \frac{n_u - n_l}{2} \rceil}}, \frac{4}{T_0 S^2 } )} $$ where $S$ is the size of the validation dataset and $A > 0$ is a constant.
\end{lemma}


While the above lemma provides a worst-case bound on the number of iterations required to find an optimal solution, it is practically infeasible to achieve the large bound.
\citet{Ben-Ameur2004} proposes an algorithm for computing $T_0$ and shows that SA performs well experimentally with fewer iterations than estimated by \citet{mitra1986convergence}. We use the temperature initialization algorithm from \cite{Ben-Ameur2004} to estimate the initial temperature $T_0$. At a high level, this algorithm back-computes an initial value for the temperature for which the expected value of acceptance probabilities for positive transitions from a random initial distribution is greater than some predefined threshold. 


\paragraph{Stopping Criterion.} We use time as a stopping criterion for our SA runs i.e., we run the SA search for time less than some hyperparameter $time\_limit$.

\begin{algorithm}

\DontPrintSemicolon
\KwIn{Unfair neural network $\mathcal{M}$, Penalty multiplier $p$, Threshold multiplier $t$, Minimum and maximum number of neurons to drop $[n_l, n_u]$, Algorithm Type $alg\_type$, Time Limit $time\_limit$}
\KwOut{Repaired neural network $\mathcal{M}_\star$, Desirable state $s_\star$, Best cost $cost_\star$}

$s$ $\gets$ \texttt{random\_state}($\mathcal{M}, n_l, n_u$)

$s_\star$, $start\_time$ $\gets$ $\phi$, curr\_time()

$cost$ $\gets$ \texttt{compute\_cost}($\mathcal{M}, s, p, t$)

$cost_\star$ $\gets$ \texttt{compute\_cost}($\mathcal{M}, s_\star, p, t$)

$T_0$ $\gets$ \texttt{estimate\_temperature}($\mathcal{M}, s$)

\While{$curr\_time() - start\_time \le time\_limit$}{

    $T \gets $ \texttt{update\_temperature}($T_0$, $curr\_time()$)
    
    $s_i \gets $ \texttt{generate\_state}($s, n_l, n_u$)

    $cost_i$ $\gets$ \texttt{compute\_cost}($\mathcal{M}, s_i, p, t$)

    $\Delta E \gets cost_i - cost$
    
    \If{$\Delta E \le 0$}{
        $cost \gets cost_i$
        
        $s \gets s_i$
    }\ElseIf{$(alg\_type == \texttt{RW}) \; \lor \; (alg\_type == \texttt{SA} \; \land \; e^{-\Delta E / T} \ge \text{Uniform(0,1)})$}   
    {
        $cost \gets cost_i$
        
        $s \gets s_i$
    }
    
    \If{$cost \le cost_\star$}{
        $cost_\star \gets cost_i$
            
        $s_\star \gets s_i$
    }
}

$\mathcal{M}_\star \gets \mathcal{M} \setminus s_\star$

\textbf{return} $\mathcal{M}_\star$, $s_\star$, $cost_\star$

\caption{\neuronrepair~to mitigate unfairness in trained neural networks}
\label{alg:neuron_repair}
\end{algorithm}

\subsection{Random Walk Search}

The Random Walk (RW) strategy samples a random state $s \in \mathcal{S}$, and recursively samples states from the neighborhood  $\Gamma(\cdot)$ to explore the search space. The RW strategy then records the best state discovered throughout the walk. We use the same cost function as SA in RW to determine the desirable state.  A key difference between RW and SA search is that in RW, we always transition to a new state, regardless of cost. In contrast, in SA, we always transition to a new state with a lower cost and transition with some probability if it has a higher cost. This is highlighted in Line 14 in Algorithm~\ref{alg:neuron_repair}. A RW can also be considered an SA run with infinite temperature, i.e., the transition probability is always 1.

\section{Experiments}
\label{sec:expr}
We pose the following research questions:

\begin{enumerate}[start=1,label={\bfseries RQ\arabic*},leftmargin=3em]

\item How successful are randomized algorithms in repairing the unfairness of DNNs via dropouts?

\vspace{0.25 em}
\item Are there dropout strategies that improve fairness and utility together?

\vspace{0.25 em}
\item What are the design considerations of search algorithms for efficient and effective fairness
repair of DNNs via dropout?

\vspace{0.25 em}
\item How do dropout strategies compare to the state-of-the-art post-processing (bias) mitigators?

\end{enumerate}

\subsection{Datasets and Models}

We evaluate \neuronrepair~with five different datasets from fairness literature. For two of the datasets, we consider two different protected groups, which effectively results in a total of 7 benchmarks. Table~\ref{table:dataset} presents an overview of the datasets and protected groups. The \textit{Adult} Census Income~\cite{Dua:2019-census}, \textit{Bank} Marketing~\cite{Dua:2019-bank}, \textit{Compas} Software~\cite{compas-dataset}, \textit{Default} Credit ~\cite{Dua:2019-credit}, and Medical Expenditure (\textit{MEPS16}) \cite{ahrqMedicalExpenditure} are binary classification tasks to predict whether an individual has income over $50$K, is likely to subscribe, has a low reoffending risk, is likely to default on the credit card payment, and is likely to utilize medical benefits, respectively.

We used DNNs trained over the dataset benchmarks as the machine learning model.
Table~\ref{t:model_arch} highlights the architectures of the DNN models used for each dataset. We use ReLU as the activation function after the linear layers. For the \textit{Bank}, \textit{Default}, and \textit{Compas} datasets we use a dropout of 0.2 during training. For the \textit{Adult} and \textit{MEPS16} datasets, we set the dropout to 0.1. For the \textit{Compas} dataset, we use Adam as the optimizer with a learning rate of $0.001$. For the other datasets, we use SGD with a learning rate of $0.01$. For data preprocessing, we utilize standard techniques such as one-hot encoding for categorical features followed by min-max or standard scaling for numerical features. For the \textit{Compas} dataset, we use a version of the dataset used in \cite{tizpaz2022fairness} that has 12 features after feature selection. The modified \textit{Compas} dataset is available online\footnote{https://github.com/Tizpaz/Parfait-ML/blob/main/subjects/datasets/compas}.


\begin{table}[]
\caption{Architectures of the DNNs used for the datasets.}
\label{t:model_arch}
\begin{tabular}{|c|c|}
\hline
\textbf{Dataset}             & \textbf{\begin{tabular}[c]{@{}c@{}}Model Architecture\\ $[L_I, L_0, \ldots, L_n, L_O]$\end{tabular}} \\ \hline
Adult Census Income          & [34, 64, 128, 64, 1]                                                                                 \\ \hline
Compas Software              & [12, 32, 32, 1]                                                                                      \\ \hline
Bank Marketing               & [32, 32, 32, 1]                                                                                      \\ \hline
Default Credit               & [30, 16, 16, 16, 1]                                                                                  \\ \hline
Medical Expenditure (MEPS16) & [138, 128, 128, 128, 1]                                                                              \\ \hline
\end{tabular}
\end{table}

\subsection{Technical Details}

We run all our experiments on a desktop running Ubuntu 22.04.3 LTS with an Intel(R) Core(TM) i7-7700 CPU @ \SI{3.60}{GHz} processor, \SI{32}{GB} RAM, and a \SI{1}{TB} HDD. The neural networks and repair algorithms were implemented using \texttt{python3.8}, \texttt{torch==2.0.1}, \\ \texttt{numpy==1.24.3}, and \texttt{scikit-learn==1.3.1}.

\subsection{Experimental Setup}

For each dataset, we evaluate the performance of the \neuronrepair~algorithms using $10$ different random seeds. The seeds are set for the \texttt{torch}, \texttt{numpy}, and \texttt{scikit-learn} libraries before DNN training. Specifically, the seed determines the randomness of the training, validation, and test splits and the randomness of model training, such as sampling batches and initializing the weights of the DNN. The \neuronrepair~algorithms themselves are not seeded. Each SA and RW run is unique. 

During training, the training dataset was used for gradient descent, and the validation dataset was used for hyperparameter tuning. We use the model from the epoch with the highest validation F1 score. The validation dataset is also used by \neuronrepair~to identify the desirable set of neurons to drop. We evaluate the fairness of the original and fixed (repaired) models using the test dataset. 

Unless otherwise specified, all runs reported in the paper have a time-out limit of $1$ hour with the following cost function:

\begin{equation*}
      C(s) = EOD_s + 3.0 \cdot EOD_{s_0} \cdot \mathds{1} (F1_{s} < (0.98 \cdot F1_{s_0})).
\end{equation*}

The F1 threshold is 98\% of the validation F1 score of the unfair DNN, i.e., we tolerate a 2\% degradation in the F1 score to improve fairness. A penalty of $3\times$ the baseline unfairness is added to the fairness of the current state if its F1 score is less than the threshold. For all datasets, we set the minimum number of neurons to $n_l=2$ and vary the maximum number of neurons $n_u$ between $20\%-40\%$ of the number of hidden layer neurons. The $n_u$ values are 50, 24, 24, 20, and 135 for \textit{Adult}, \textit{Compas}, \textit{Bank}, \textit{Default}, and \textit{MEPS16} respectively. For all datasets, we use a train/validation/test split of 60\%/20\%/20\%. We set the threshold for the acceptance probabilities in the temperature initialization algorithm \cite{Ben-Ameur2004} to $0.75$.

\begin{table*}[!t]
\caption{Datasets used in our experiments.}
\label{table:dataset}
\centering
\resizebox{0.7\textwidth}{!}{
\begin{tabu}{|l|l|l|ll|ll|}
  \hline
  \multirow{2}{*}{\textbf{Dataset}} & \multirow{2}{*}{\textbf{|Instances|}} & \multirow{2}{*}{\textbf{|Features|}} & \multicolumn{2}{c|}{\textbf{Protected Groups}} & \multicolumn{2}{c|}{\textbf{Outcome Label}} \\
   &  & & \textit{Group1} & \textit{Group2} & \textit{Label 1} & \textit{Label 0}  \\
  \hline
  \textit{Adult} Census \cite{Dua:2019-census} & \multirow{2}{*}{$48,842$} & \multirow{2}{*}{$14$} & Sex-Male & Sex-Female & \multirow{2}{*}{High Income} & \multirow{2}{*}{Low Income} \\ \cline{4-5}
  Income &  &    &  Race-White & Race-Non White  &   &    \\
  \hline
  \multirow{2}{*}{\textit{Compas} Software} \cite{compas-dataset} & \multirow{2}{*}{$7,214$} & \multirow{2}{*}{$28$} & Sex-Male & Sex-Female & \multirow{2}{*}{Did not Reoffend} & \multirow{2}{*}{Reoffend} \\  \cline{4-5}
  &  &    &  Race-Caucasian & Race-Non Caucasian  &   &    \\
  \hline
  \textit{Bank} Marketing \cite{Dua:2019-bank}  & $45,211$ & $17$ & Age-Young & Age-Old & Subscriber & Non-subscriber \\
  \hline
  \textit{Default} Credit \cite{Dua:2019-credit}  & $13,636$ & $23$ & Sex-Male & Sex-Female & Default & Not Default \\
  \hline
  Medical Expenditure (\textit{MEPS16}) \cite{ahrqMedicalExpenditure}  & $15,675$ & $138$ & Race-White & Race-Non White & Utilized Benefits & Not Utilized Benefits \\
  \hline
  
\end{tabu}
}
\end{table*}

\begin{table*}[]
\caption{Fairness and model utility metrics of the unrepaired (original) DNN.}
\label{t:unrepaired_results}
\centering
\resizebox{0.6\textwidth}{!}{
\begin{tabular}{|c|cccccc|}
\hline
\multirow{4}{*}{\textbf{Dataset}} & \multicolumn{6}{c|}{\textbf{Unfair Neural Network}}                                                                                                                                                                                                                                                      \\ \cline{2-7} 
                                  & \multicolumn{3}{c|}{\multirow{2}{*}{Validation}}                                                                                                              & \multicolumn{3}{c|}{\multirow{2}{*}{Test}}                                                                                               \\
                                  & \multicolumn{3}{c|}{}                                                                                                                                         & \multicolumn{3}{c|}{}                                                                                                                    \\ \cline{2-7} 
                                  & \multicolumn{1}{c|}{\textit{EOD}}         & \multicolumn{1}{c|}{\textit{F1}}                        & \multicolumn{1}{c|}{\textit{Accuracy}}                  & \multicolumn{1}{c|}{\textit{EOD}}         & \multicolumn{1}{c|}{\textit{F1}}                        & \textit{Accuracy}                  \\ \hline
Adult (Sex)                       & \multicolumn{1}{c|}{$9.704\% \pm 1.405$}  & \multicolumn{1}{c|}{\multirow{2}{*}{$0.68 \pm 0.005$}}  & \multicolumn{1}{c|}{\multirow{2}{*}{$0.857 \pm 0.003$}} & \multicolumn{1}{c|}{$11.639\% \pm 2.326$} & \multicolumn{1}{c|}{\multirow{2}{*}{$0.667 \pm 0.008$}} & \multirow{2}{*}{$0.851 \pm 0.003$} \\ \cline{1-2} \cline{5-5}
Adult (Race)                      & \multicolumn{1}{c|}{$8.352\% \pm 2.806$}  & \multicolumn{1}{c|}{}                                   & \multicolumn{1}{c|}{}                                   & \multicolumn{1}{c|}{$8.251\% \pm 3.195$}  & \multicolumn{1}{c|}{}                                   &                                    \\ \hline
COMPAS (Sex)                      & \multicolumn{1}{c|}{$2.307\% \pm 0.679$}  & \multicolumn{1}{c|}{\multirow{2}{*}{$0.968 \pm 0.002$}} & \multicolumn{1}{c|}{\multirow{2}{*}{$0.97 \pm 0.002$}}  & \multicolumn{1}{c|}{$2.522\% \pm 0.817$}  & \multicolumn{1}{c|}{\multirow{2}{*}{$0.967 \pm 0.004$}} & \multirow{2}{*}{$0.969 \pm 0.004$} \\ \cline{1-2} \cline{5-5}
COMPAS (Race)                     & \multicolumn{1}{c|}{$2.022\% \pm 0.884$}  & \multicolumn{1}{c|}{}                                   & \multicolumn{1}{c|}{}                                   & \multicolumn{1}{c|}{$2.96\% \pm 1.088$}   & \multicolumn{1}{c|}{}                                   &                                    \\ \hline
Bank                              & \multicolumn{1}{c|}{$13.752\% \pm 2.567$} & \multicolumn{1}{c|}{$0.553 \pm 0.006$}                  & \multicolumn{1}{c|}{$0.842 \pm 0.003$}                  & \multicolumn{1}{c|}{$14.665\% \pm 2.114$} & \multicolumn{1}{c|}{$0.553 \pm 0.004$}                  & $0.84 \pm 0.003$                   \\ \hline
Default                           & \multicolumn{1}{c|}{$9.068\% \pm 1.782$}  & \multicolumn{1}{c|}{$0.538 \pm 0.006$}                  & \multicolumn{1}{c|}{$0.774 \pm 0.007$}                  & \multicolumn{1}{c|}{$8.962\% \pm 1.772$}  & \multicolumn{1}{c|}{$0.53 \pm 0.006$}                   & $0.769 \pm 0.007$                  \\ \hline
MEPS16                            & \multicolumn{1}{c|}{$20.167\% \pm 2.32$}  & \multicolumn{1}{c|}{$0.543 \pm 0.009$}                  & \multicolumn{1}{c|}{$0.79 \pm 0.005$}                   & \multicolumn{1}{c|}{$20.641\% \pm 2.527$} & \multicolumn{1}{c|}{$0.533 \pm 0.01$}                   & $0.788 \pm 0.009$                  \\ \hline
\end{tabular}
}
\end{table*}

\begin{table*}[]
\caption{Fairness and model utility metrics of the repaired DNN.}
\label{t:repaired_results}
\resizebox{\textwidth}{!}{
\centering
\begin{tabular}{|c|cccccccccccc|}
\hline
\multirow{4}{*}{\textbf{Dataset}} & \multicolumn{12}{c|}{\textbf{Repaired Neural Network}}                                                                                                                                                                                                                                                                                                                                                                                                                                       \\ \cline{2-13} 
                                  & \multicolumn{6}{c|}{Simulated Annealing}                                                                                                                                                                                                                & \multicolumn{6}{c|}{Random Walk}                                                                                                                                                                                                   \\ \cline{2-13} 
                                  & \multicolumn{3}{c|}{Validation}                                                                                            & \multicolumn{3}{c|}{Test}                                                                                                  & \multicolumn{3}{c|}{Validation}                                                                                            & \multicolumn{3}{c|}{Test}                                                                             \\ \cline{2-13} 
                                  & \multicolumn{1}{c|}{\textit{EOD}}        & \multicolumn{1}{c|}{\textit{F1}}       & \multicolumn{1}{c|}{\textit{Accuracy}} & \multicolumn{1}{c|}{\textit{EOD}}        & \multicolumn{1}{c|}{\textit{F1}}       & \multicolumn{1}{c|}{\textit{Accuracy}} & \multicolumn{1}{c|}{\textit{EOD}}        & \multicolumn{1}{c|}{\textit{F1}}       & \multicolumn{1}{c|}{\textit{Accuracy}} & \multicolumn{1}{c|}{\textit{EOD}}        & \multicolumn{1}{c|}{\textit{F1}}       & \textit{Accuracy} \\ \hline
Adult (Sex)                       & \multicolumn{1}{c|}{$5.618\% \pm 0.807$} & \multicolumn{1}{c|}{$0.668 \pm 0.005$} & \multicolumn{1}{c|}{$0.856 \pm 0.004$} & \multicolumn{1}{c|}{$7.259\% \pm 1.697$} & \multicolumn{1}{c|}{$0.652 \pm 0.01$}  & \multicolumn{1}{c|}{$0.849 \pm 0.004$} & \multicolumn{1}{c|}{$6.01\% \pm 0.704$}  & \multicolumn{1}{c|}{$0.667 \pm 0.005$} & \multicolumn{1}{c|}{$0.855 \pm 0.003$} & \multicolumn{1}{c|}{$7.358\% \pm 1.063$} & \multicolumn{1}{c|}{$0.652 \pm 0.01$}  & $0.849 \pm 0.004$ \\ \hline
Adult (Race)                      & \multicolumn{1}{c|}{$3.298\% \pm 2.069$} & \multicolumn{1}{c|}{$0.667 \pm 0.005$} & \multicolumn{1}{c|}{$0.854 \pm 0.003$} & \multicolumn{1}{c|}{$4.976\% \pm 1.816$} & \multicolumn{1}{c|}{$0.656 \pm 0.008$} & \multicolumn{1}{c|}{$0.849 \pm 0.004$} & \multicolumn{1}{c|}{$3.965\% \pm 1.665$} & \multicolumn{1}{c|}{$0.667 \pm 0.005$} & \multicolumn{1}{c|}{$0.853 \pm 0.003$} & \multicolumn{1}{c|}{$4.785\% \pm 2.085$} & \multicolumn{1}{c|}{$0.658 \pm 0.009$} & $0.849 \pm 0.003$ \\ \hline
COMPAS (Sex)                      & \multicolumn{1}{c|}{$0.468\% \pm 0.548$} & \multicolumn{1}{c|}{$0.955 \pm 0.004$} & \multicolumn{1}{c|}{$0.959 \pm 0.003$} & \multicolumn{1}{c|}{$2.921\% \pm 1.446$} & \multicolumn{1}{c|}{$0.954 \pm 0.08$}  & \multicolumn{1}{c|}{$0.957 \pm 0.008$} & \multicolumn{1}{c|}{$0.54\% \pm 0.535$}  & \multicolumn{1}{c|}{$0.956 \pm 0.005$} & \multicolumn{1}{c|}{$0.959 \pm 0.004$} & \multicolumn{1}{c|}{$2.233\% \pm 1.022$} & \multicolumn{1}{c|}{$0.954 \pm 0.007$} & $0.957 \pm 0.006$ \\ \hline
COMPAS (Race)                     & \multicolumn{1}{c|}{$0.56\% \pm 0.71$} & \multicolumn{1}{c|}{$0.955 \pm 0.002$} & \multicolumn{1}{c|}{$0.959 \pm 0.002$} & \multicolumn{1}{c|}{$2.239\% \pm 1.003$}  & \multicolumn{1}{c|}{$0.954 \pm 0.005$} & \multicolumn{1}{c|}{$0.957 \pm 0.004$} & \multicolumn{1}{c|}{$0.58\% \pm 0.715$}  & \multicolumn{1}{c|}{$0.956 \pm 0.003$} & \multicolumn{1}{c|}{$0.959 \pm 0.002$} & \multicolumn{1}{c|}{$2.159\% \pm 1.07$} & \multicolumn{1}{c|}{$0.955 \pm 0.004$} & $0.958 \pm 0.004$ \\ \hline
Bank                              & \multicolumn{1}{c|}{$0.871\% \pm 0.509$} & \multicolumn{1}{c|}{$0.547 \pm 0.007$} & \multicolumn{1}{c|}{$0.892 \pm 0.01$}  & \multicolumn{1}{c|}{$7.257\% \pm 3.533$} & \multicolumn{1}{c|}{$0.537 \pm 0.014$} & \multicolumn{1}{c|}{$0.888 \pm 0.01$}  & \multicolumn{1}{c|}{$1.714\% \pm 0.921$} & \multicolumn{1}{c|}{$0.548 \pm 0.006$} & \multicolumn{1}{c|}{$0.882 \pm 0.01$}  & \multicolumn{1}{c|}{$7.595\% \pm 2.733$} & \multicolumn{1}{c|}{$0.548 \pm 0.008$} & $0.881 \pm 0.008$ \\ \hline
Default                           & \multicolumn{1}{c|}{$1.14\% \pm 0.98$}   & \multicolumn{1}{c|}{$0.529 \pm 0.007$} & \multicolumn{1}{c|}{$0.794 \pm 0.016$} & \multicolumn{1}{c|}{$2.749\% \pm 0.827$} & \multicolumn{1}{c|}{$0.519 \pm 0.006$} & \multicolumn{1}{c|}{$0.79 \pm 0.015$}  & \multicolumn{1}{c|}{$2.045\% \pm 1.251$} & \multicolumn{1}{c|}{$0.53 \pm 0.007$}  & \multicolumn{1}{c|}{$0.794 \pm 0.015$} & \multicolumn{1}{c|}{$3.124\% \pm 0.937$} & \multicolumn{1}{c|}{$0.523 \pm 0.005$} & $0.79 \pm 0.015$  \\ \hline
MEPS16                            & \multicolumn{1}{c|}{$4.589\% \pm 1.294$} & \multicolumn{1}{c|}{$0.535 \pm 0.01$}  & \multicolumn{1}{c|}{$0.86 \pm 0.006$}  & \multicolumn{1}{c|}{$8.426\% \pm 2.311$} & \multicolumn{1}{c|}{$0.507 \pm 0.02$}  & \multicolumn{1}{c|}{$0.853 \pm 0.005$} & \multicolumn{1}{c|}{$6.622\% \pm 1.183$} & \multicolumn{1}{c|}{$0.535 \pm 0.01$}  & \multicolumn{1}{c|}{$0.856 \pm 0.005$} & \multicolumn{1}{c|}{$9.86\% \pm 2.623$}  & \multicolumn{1}{c|}{$0.513 \pm 0.018$} & $0.851 \pm 0.007$ \\ \hline
\end{tabular}
}
\end{table*}

\begin{figure*}
    \centering
    \includegraphics[scale=0.17]
    {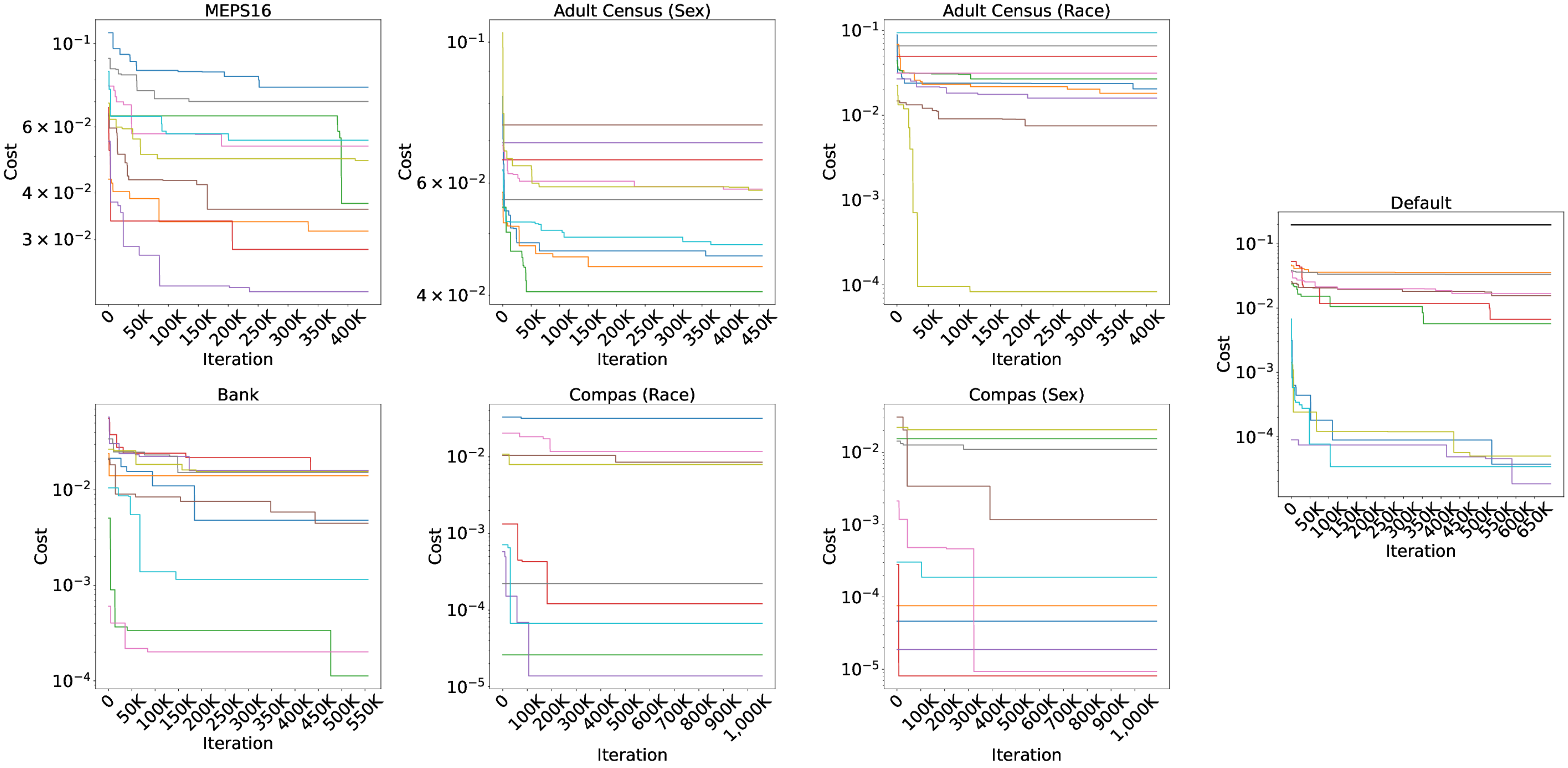}
    \caption{Evolution of the cost of the best state (logarithmic scale) found during the search on validation dataset using Simulated Annealing (SA) with 7 benchmarks and 10 seeds each.}
    \label{fig:sa_graph}
\end{figure*}

\begin{figure*}
    \centering
    \includegraphics[scale=0.17]
    {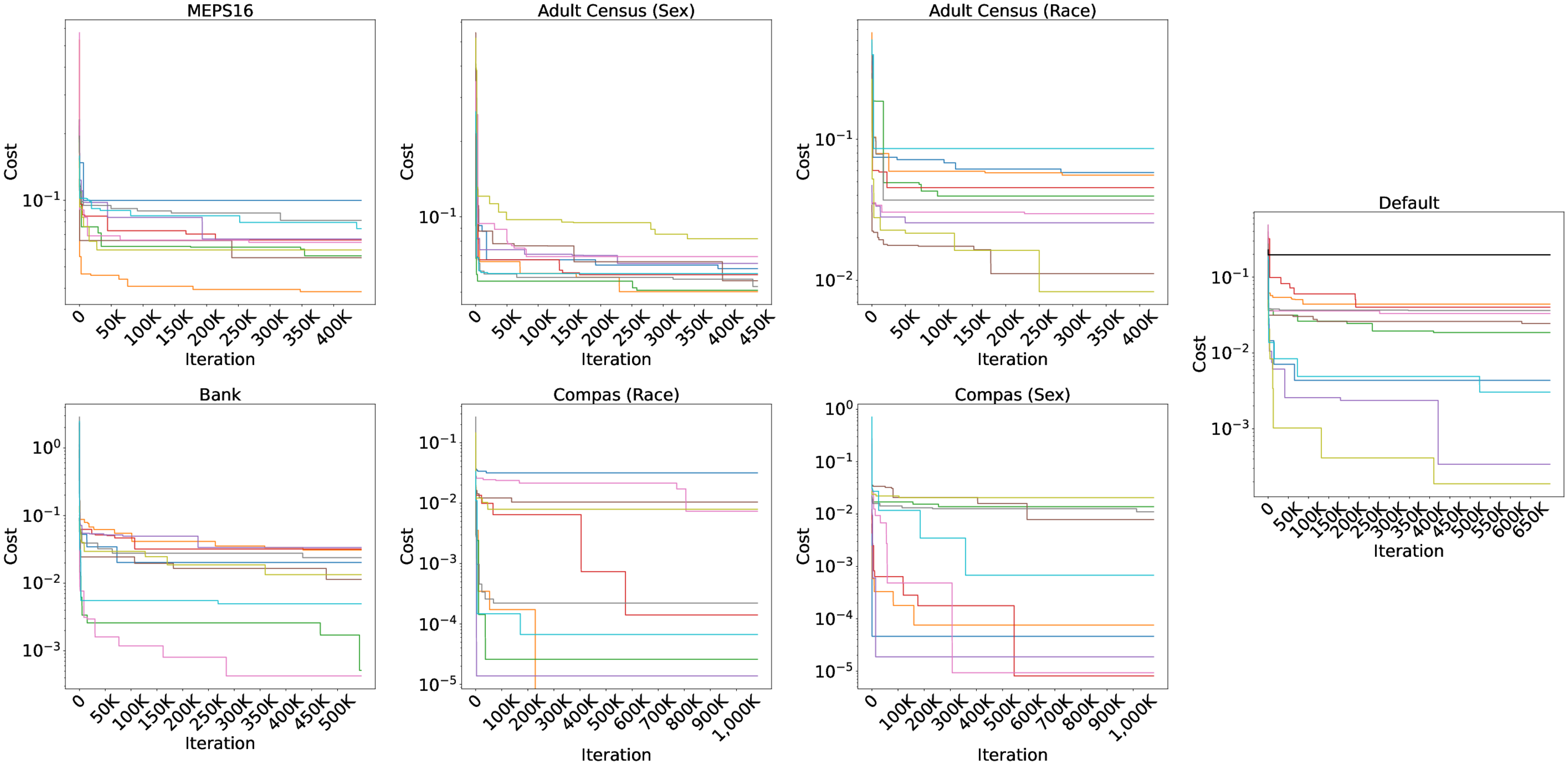}
    \caption{Evolution of the cost of the best state (logarithmic scale) found during the search on validation dataset using Random Walk (RW) with 7 benchmarks and 10 seeds each.}
    \label{fig:rw_graph}
\end{figure*}

\subsubsection{Effectiveness of randomized algorithms on mitigating unfairness using dropout (RQ1)}

The first research question is answered by analyzing the data in Figure~\ref{fig:sa_graph}, Figure~\ref{fig:rw_graph}, Table~\ref{t:unrepaired_results}, and Table~\ref{t:repaired_results}. 

Figure~\ref{fig:sa_graph} and Figure~\ref{fig:rw_graph} highlight the cost of the best state found for the SA and RW over time, respectively, for 10 different seeds per data subject using the \emph{logarithmic scale}. In all cases, we see a positive trend where the cost improves over time, and the search identifies better subsets of neurons to drop. We note the seemingly flat plots correspond to those cases where a good state was found in the beginning. The exception is the black line in the sub-plot for the \textit{Default} dataset in both figures as it corresponds to a failed run; i.e., the best state found has an F1 score lesser than $0.98\times$ the F1 score of the initial state. Out of 70 runs (10 per benchmark), the SA and RW found a desirable state with $0.98\times$ the initial F1 score in 69 cases. 
In addition, we re-run the failed \textit{Default} experiment with an F1 threshold multiplier of $0.96$ and identify a desirable state with a test set fairness of $6.26\%$, an improvement from $13.2\%$. The progress graphs of the RW strategy are ``steep'' initially as the initial states are unfair and are heavily penalized by the penalty multiplier as RW generates intermediate states with an F1 score less than the threshold. However, due to the cost function and probabilistic transitions, SA only explores worse states with a low probability. Thus, the progress graphs are not as steep initially. 

Table~\ref{t:unrepaired_results} and Table~\ref{t:repaired_results} highlight the mean and $95\%$ confidence intervals for the F1 score, accuracy, and EOD for all data subjects aggregated across 10 seeds. We observe that in all cases, both \neuronrepair~algorithms improve the EOD. However, SA outperformed RW in all cases for the validation set while ensuring that the F1 score reduces by at most $2\%$. We can conclude that SA is the more effective algorithm in identifying the desirable state during the search procedure, subject to the F1 threshold. 
On the test set, we observe that both \neuronrepair~algorithms improve the EOD. SA outperforms RW for 4 datasets (\textit{Adult (Sex)}, \textit{Bank}, \textit{Default}, and \textit{MEPS16}).  We observe the largest absolute improvement of $12.215\%$ in EOD ($20.641\%$ to $8.426\%$ ) for the \textit{MEPS16} dataset using SA. The largest relative improvement of $69.43\%$ ($8.962\%$ to $2.749\%$)  was observed in the \textit{Default} dataset using SA. However, SA
did not improve the EOD on the \textit{COMPAS (Sex)}, which is a pathological case.


\begin{table}[h]
\caption{Fairness metrics for COMPAS (Sex)}
\resizebox{0.49\textwidth}{!}{
\begin{tabular}{|c|cccccccccc|}
\hline
\textbf{Model} & \multicolumn{10}{c|}{\textbf{EOD for 10 seeds}}                                                                                                                                                                                                                  \\ \hline
Unfair         & \multicolumn{1}{c|}{2.2\%}  & \multicolumn{1}{c|}{2.5\%}  & \multicolumn{1}{c|}{2.35\%} & \multicolumn{1}{c|}{1.87\%} & \multicolumn{1}{c|}{4.1\%}  & \multicolumn{1}{c|}{2.15\%} & \multicolumn{1}{c|}{1.38\%} & \multicolumn{1}{c|}{4.26\%} & \multicolumn{1}{c|}{0.79\%} & 3.6\%  \\ \hline
Repaired (SA)  & \multicolumn{1}{c|}{1.97\%} & \multicolumn{1}{c|}{4.28\%} & \multicolumn{1}{c|}{2.35\%} & \multicolumn{1}{c|}{4.58\%} & \multicolumn{1}{c|}{4.37\%} & \multicolumn{1}{c|}{0.12\%} & \multicolumn{1}{c|}{0.47\%} & \multicolumn{1}{c|}{4.83\%} & \multicolumn{1}{c|}{5.5\%}  & 0.75\% \\ \hline
Repaired (RW)  & \multicolumn{1}{c|}{1.25\%} & \multicolumn{1}{c|}{3.61\%} & \multicolumn{1}{c|}{3.19\%} & \multicolumn{1}{c|}{2.5\%}  & \multicolumn{1}{c|}{2.4\%}  & \multicolumn{1}{c|}{0.53\%} & \multicolumn{1}{c|}{0.28\%} & \multicolumn{1}{c|}{4.54\%} & \multicolumn{1}{c|}{3.07\%} & 0.92\% \\ \hline
\end{tabular}
}
\label{t:compas_sex}
\end{table}

\vspace{0.25 em}
\noindent \textit{Pathological Case.} Table~\ref{t:compas_sex} highlights the individual EOD scores for all seeds of \textit{COMPAS (Sex)} with the original and repaired networks. Here, we observe that SA improves the EOD in 4 cases, does not improve in 1 case, and has a worse EOD in 5 cases. RW improves the EOD in 5 cases and has a worse EOD in the other 5 cases.
There are two reasons for the different trends in \textit{COMPAS (Sex)}. First, there is a discrepancy in the original model statistics between the validation and test splits. The difference arises because the DNN tends to overfit slightly on the training and validation sets. This is evident from model metrics in Table~\ref{t:unrepaired_results}. Second, the \textit{COMPAS} datasets by default are quite fair (according to EOD) compared to other datasets as their unfairness is around 2\%. During the search, the SA algorithm correctly identifies a desirable state according to the validation dataset and performs better than RW. However, given the low initial unfairness and the model's tendency to perform better on the training and validation sets, we observe minor discrepancies in the test set. The average unfairness, however, increases by 0.399\% (2.522\% to 2.921\%), which is negligible.  For both \neuronrepair~algorithms, the validation fairness improves in the \textit{COMPAS (Sex)} experiment.

\begin{tcolorbox}[boxrule=1pt,left=1pt,right=1pt,top=1pt,bottom=1pt]
\textbf{Answer RQ1:} The randomized algorithms effectively mitigate unfairness via dropouts by de-activating a desirable subset of neurons. On the test set, Fairness improves by up to 69\%. The SA algorithm performs better than the RW algorithm.
\end{tcolorbox}

\subsubsection{Dropout strategies improving both fairness and utility (RQ2)}

The second research question is answered by observing the F1 score and accuracy (model utility metrics), besides the EOD score (fairness) in Table~\ref{t:unrepaired_results} vs. Table~\ref{t:repaired_results}. For all datasets, the F1 score decreases for the validation and test sets. The decrease in F1 score is accompanied by an improvement in fairness which highlights the tradeoff between the model utility and fairness. However, we observe that the accuracy increases for the \textit{Bank}, \textit{Default}, and \textit{MEPS16} datasets. The biggest improvement in accuracy is observed in MEPS16, where the validation and test accuracy increase from 0.79 and 0.788 to 0.86 and 0.853 for SA and to 0.856 to 0.851 for RW. For \textit{Default}, the validation and test accuracies increase from 0.774 and 0.769 to 0.794 and 0.79 for both SA and RW. For Bank, the validation and test accuracies increase from 0.842 and 0.84 to 0.892 and 0.88 for SA and 0.882 and 0.881 for RW. 

The opposing trends of F1 score and accuracy can be attributed to the increase in negative predictions (0 is the negative class and 1 is positive) as neurons continue to be dropped out from the DNN. The increase in negative predictions favors the true negatives as the datasets are highly imbalanced with more negative samples. The \textit{Default}, \textit{Bank}, and \textit{MEPS16} datasets have 78\%, 88\%, and 83\% of the data belonging to the negative class. The trained DNN models were optimized for F1 score on the validation dataset to improve performance on the underrepresented positive class. The accuracy increases as the increase in true negatives far outweighs the drop in true positives. The F1 score, however, always decreases as we lose precision and recall when the true positives decrease and the false negatives increase.

\begin{tcolorbox}[boxrule=1pt,left=1pt,right=1pt,top=1pt,bottom=1pt]
\textbf{Answer RQ2:} The overall accuracy, as a model utility metric, may improve along with fairness using dropout strategies, as accuracy does not account for false positives and false negatives in imbalanced datasets. However, the F1 score model utility metric decreases as fairness increases.

\end{tcolorbox}

\subsubsection{Hyperparameters of randomized algorithms and fairness (RQ3)}

The \neuronrepair~algorithms have 4 hyperparameters: F1 threshold multiplier, the minimum and maximum number of neurons to drop, time-out limit,
and F1 penalty multiplier. 

\begin{figure}
    \centering
    \includegraphics[scale=0.4]{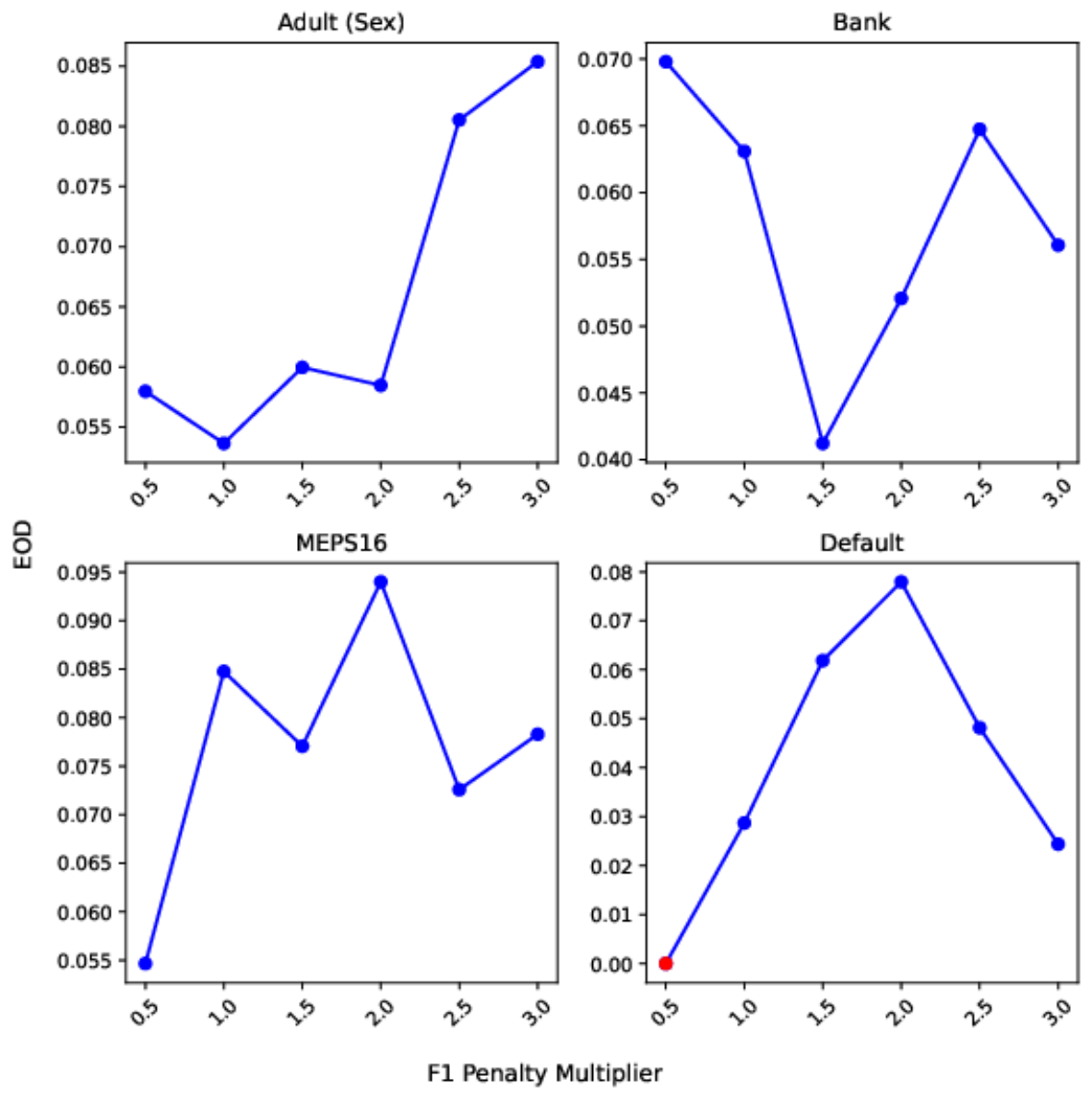}
    \caption{Effect of F1 penalty multiplier on the EOD}
    \label{fig:f1_pen_var}
\end{figure}

\begin{itemize}[leftmargin=0.3cm]
    \item The F1 threshold multiplier is inversely proportional to fairness improvement. For example, as highlighted in RQ1, the fairness of the Default experiment improves when the F1 threshold is reduced from 0.98 to 0.96. 
    \item Decreasing the minimum number of neurons and increasing the maximum number of neurons can have a positive effect on fairness, provided the time-out limit increases. Consider two ranges $[n_{l1}, n_{u1}]$ and $[n_{l2}, n_{u2}]$ such that $n_{l1} < n_{l2} < n_{u2} < n_{u1}$. The search space of $[n_{l2}, n_{u2}]$ is then a sub-space of $[n_{l1}, n_{u1}]$. 
    \item The time-out positively affects fairness as the \neuronrepair~algorithms have more time to explore the search space. 
    \item The F1 penalty multiplier has a more nuanced effect on the unfairness as it controls the exploration vs. exploitation trade-offs.
    A low penalty multiplier increases the probability of state transitions when the F1 score is less than our threshold, increasing the SA run's randomness to explore more of the search space outside of the current best state. A high penalty multiplier would decrease the probability of state transitions, thereby encouraging the SA algorithm to exploit the search space near the current best state. The penalty multiplier \textit{p} directly affects the cost difference $\Delta E$ between states with acceptable and unacceptable F1 scores. From Line 14 in Algorithm~\ref{alg:neuron_repair}, we can see that a higher cost difference results in a lower transition probability.
    For a fixed temperature, a higher \textit{p} would increase the cost difference between good and bad states, thereby decreasing the probability of transitioning and encouraging exploitation. A lower \textit{p} would result in a lower cost difference and increase the transition probability, thereby encouraging exploration. To determine the effect of the penalty multiplier, we run experiments with the \textit{Adult}, \textit{MEPS16}, \textit{Default}, and \textit{Bank} datasets with one seed by varying the F1 penalty multiplier $p \in \{0.5, 1.0, 1.5, 2.0, 2.5, 3.0\}$ and F1 threshold multiplier to 0.98. Figure~\ref{fig:f1_pen_var} shows the results. The red dot in the \textit{Default} dataset corresponds to a run in which SA failed to find a desirable state within the F1 threshold. We can see that \textit{Adult} dataset performs better on lower F1 penalties. The \textit{Bank} dataset has a ``sweet'' spot in between while \textit{Default} performs worse on intermediate F1 penalties. The \textit{MEPS16} does not follow a clear trend. The results show that the F1 penalty multiplier is a hyperparameter that trades off explorations vs. exploitation and  
    requires tuning for each dataset and model. 
\end{itemize}

\begin{tcolorbox}[boxrule=1pt,left=1pt,right=1pt,top=1pt,bottom=1pt]
\textbf{Answer RQ3:}
The fairness of the repaired model can improve as the F1 threshold decreases, the number of neurons to drop increases, and the time-out limit increases for both \neuronrepair-SA and \neuronrepair-RW algorithms. The F1 penalty multiplier controls a trade-off between explorations
vs. exploitation, which requires tuning for each specific benchmark. A higher F1 penalty multiplier encourages SA to exploit the local solutions, whereas a lower penalty encourages exploring the global space.
\end{tcolorbox}

\subsubsection{Comparing to the state-of-the-art (RQ4)}
We compare the efficacy of \toolname against a state-of-the-art post-processing bias mitigator.
\textsc{Dice}~\cite{10.1109/ICSE48619.2023.00136} uses a fault localization technique
via \text{do} logic to pinpoint a single neuron that significantly influences fairness.
The approach computes the amounts of variation in the outcomes of each layer that depends on
the sensitive attributes and de-activates a single neuron with the highest effects
on fairness. Since the search space evaluates a single neuron, we brute-force the entire search space 
of one neuron dropout in linear time. We note that the best improvement, feasible by \textsc{Dice},
cannot be more than the brute-force search.
Table~\ref{t:dice_results} highlights the test fairness after dropping a single neuron that most impacts fairness. The results show that \neuronrepair~substantially outperforms \textsc{DICE} in improving fairness. We also note a similar trend with \textit{COMPAS (Sex)}, where the unfairness increases slightly. The largest discrepancy in fairness improvement between \toolname and \textsc{Dice} is observed in the \textit{MEPS16} dataset, with an absolute difference of 10.778\% and a relative improvement of 56.124\% (19.204\% and 8.426\%) using SA.

\begin{tcolorbox}[boxrule=1pt,left=1pt,right=1pt,top=1pt,bottom=1pt]
\textbf{Answer RQ4:}
\toolname~outperforms the state-of-the-art unfairness mitigation algorithm \textsc{Dice}~\cite{10.1109/ICSE48619.2023.00136} across all benchmarks. The relative improvement of \toolname is 56\% higher than \textsc{Dice}.

\end{tcolorbox}

\begin{table}[]
\caption{Fairness and Utility of \textsc{Dice}~\cite{10.1109/ICSE48619.2023.00136}.}
\label{t:dice_results}
\resizebox{0.47\textwidth}{!}{
\begin{tabular}{|c|ccc|}
\hline
\multirow{2}{*}{\textbf{Dataset}} & \multicolumn{3}{c|}{\textbf{Test Dataset}}                                                                     \\ \cline{2-4} 
                                  & \multicolumn{1}{c|}{\textit{EOD}}         & \multicolumn{1}{c|}{\textit{F1}}       & \textit{Accuracy} \\ \hline
Adult (Sex)                       & \multicolumn{1}{c|}{$10.453\% \pm 2.266$} & \multicolumn{1}{c|}{$0.658 \pm 0.007$} & $0.851 \pm 0.003$ \\ \hline
Adult (Race)                      & \multicolumn{1}{c|}{$8.092\% \pm 3.253$}  & \multicolumn{1}{c|}{$0.662 \pm 0.008$} & $0.851 \pm 0.003$ \\ \hline
COMPAS (Sex)                      & \multicolumn{1}{c|}{$3.229\% \pm 0.774$}    & \multicolumn{1}{c|}{$0.964 \pm 0.005$} & $0.957 \pm 0.004$ \\ \hline
COMPAS (Race)                     & \multicolumn{1}{c|}{$2.964\% \pm 1.088$}    & \multicolumn{1}{c|}{$0.965 \pm 0.003$} & $0.968 \pm 0.003$ \\ \hline
Bank                              & \multicolumn{1}{c|}{$12.205\% \pm 2.731$}   & \multicolumn{1}{c|}{$0.567 \pm 0.005$} & $0.862 \pm 0.01$  \\ \hline
Default                           & \multicolumn{1}{c|}{$5.845\% \pm 1.816$}    & \multicolumn{1}{c|}{$0.528 \pm 0.004$} & $0.791 \pm 0.007$ \\ \hline
MEPS16                            & \multicolumn{1}{c|}{$19.204\% \pm 2.592$}   & \multicolumn{1}{c|}{$0.54 \pm 0.012$}  & $0.8 \pm 0.009$   \\ \hline
\end{tabular}
}
\end{table}

\section{Discussion}
\label{sec:discussion}

\begin{table}[]
\caption{Simulated Annealing compared to Brute-force search strategies.}
\label{t:sa_bf}
\resizebox{0.47\textwidth}{!}{
\begin{tabular}{|c|c|cccl|}
\hline
\multirow{2}{*}{\textbf{Seed}} & \multirow{2}{*}{\textbf{\begin{tabular}[c]{@{}c@{}}Unfair DNN\\ EOD\end{tabular}}} & \multicolumn{4}{c|}{\textbf{Repaired Neural Network EOD}}                                                    \\ \cline{3-6} 
                               &                                                                                               & \multicolumn{1}{c|}{SA} & \multicolumn{1}{c|}{Brute Force} & \multicolumn{2}{c|}{EOD Delta} \\ \hline
1                              & 14.087\%                                                                                      & \multicolumn{1}{c|}{6.441\%}             & \multicolumn{1}{c|}{6.415\%}     & \multicolumn{2}{c|}{0.026}     \\ \hline
2                              & 8.874\%                                                                                       & \multicolumn{1}{c|}{5.306\%}             & \multicolumn{1}{c|}{5.306\%}     & \multicolumn{2}{c|}{0.0}       \\ \hline
3                              & 8.352\%                                                                                       & \multicolumn{1}{c|}{5.755\%}             & \multicolumn{1}{c|}{5.755\%}     & \multicolumn{2}{c|}{0.0}       \\ \hline
4                              & 17.896\%                                                                                      & \multicolumn{1}{c|}{13.823\%}            & \multicolumn{1}{c|}{13.743\%}    & \multicolumn{2}{c|}{0.08}      \\ \hline
5                              & 8.062\%                                                                                       & \multicolumn{1}{c|}{5.304\%}             & \multicolumn{1}{c|}{5.25\%}      & \multicolumn{2}{c|}{0.054}     \\ \hline
\end{tabular}
}
\end{table}

\begin{table}[]
\caption{
Distribution of \textit{Best}, \textit{Good}, and \textit{Bad} states in the search space.}
\label{t:bf_state_dist}
\resizebox{0.47\textwidth}{!}{
\begin{tabular}{|c|cc|cc|cc|}
\hline
\multirow{2}{*}{\textbf{Seed}} & \multicolumn{2}{c|}{\textbf{Best State}}        & \multicolumn{2}{c|}{\textbf{Good State}}         & \multicolumn{2}{c|}{\textbf{Bad State}} \\ \cline{2-7} 
                               & \multicolumn{1}{c|}{Count} & Likelihood              & \multicolumn{1}{c|}{Count}  & Likelihood              & \multicolumn{1}{c|}{Count}      & Likelihood \\ \hline
1                              & \multicolumn{1}{c|}{1}     & $2.3\times10^{-8}$ & \multicolumn{1}{c|}{43,789} & 0.001                & \multicolumn{1}{c|}{41,669,509} & 0.967  \\ \hline
2                              & \multicolumn{1}{c|}{11}    & $2.5\times10^{-7}$ & \multicolumn{1}{c|}{6,717}  & $1.0 \times 10^{-4}$               & \multicolumn{1}{c|}{40,062,436} & 0.93  \\ \hline
3                              & \multicolumn{1}{c|}{8}     & $1.8\times10^{-7}$ & \multicolumn{1}{c|}{1,795}  & $4.1\times10^{-5}$ & \multicolumn{1}{c|}{42,997,093} & 0.998  \\ \hline
4                              & \multicolumn{1}{c|}{1}     & $2.3\times10^{-8}$ & \multicolumn{1}{c|}{790}    & $1.8\times10^{-5}$ & \multicolumn{1}{c|}{42,467,272} & 0.985  \\ \hline
5                              & \multicolumn{1}{c|}{1}     & $2.3\times10^{-8}$ & \multicolumn{1}{c|}{13,623} & $3.16\times10^{-4}$              & \multicolumn{1}{c|}{40,197,630} & 0.933  \\ \hline
\end{tabular}
}
\end{table}

\noindent
\textit{Effectiveness of Randomized Algorithms.} To understand why randomized algorithms are effective in mitigating unfairness, we compare them to brute-force strategies. We create a small DNN for the \textit{Adult (Sex)} dataset with 2 hidden layers, each with 16 neurons, to allow the brute force search to explore the entire space. We use a restricted search space with $n_l = 4$ and $n_u = 9$ that has a total of $43,076,484$ states. The time-out for SA runs is one hour. The brute force exhaustive search takes 60 hours. Table~\ref{t:sa_bf} highlights the results of both search strategies. SA finds the global optimal solution for two seeds in the search space. The EOD Delta column highlights the difference in cost of the states identified by SA and brute force. The highest cost difference of $0.054$ shows the efficacy of randomized algorithms in identifying good states.
SA is faster than brute force in reaching a good state as it follows a cost gradient to choose a path to a good state by always accepting states that improve the cost. Additionally, SA uses a probabilistic transition function that can transition to worse states and potentially escape local minima. Gradient-based methods often get stuck in local minima. Table~\ref{t:bf_state_dist} shows the distribution of the \textit{Best}, \textit{Good}, and \textit{Bad} states. The best states are the states with the global optimal cost. The cost of the good states is within 0.05 of the optimal cost. Bad states are states whose F1 scores are less than the 98\% of the baseline F1 score. We observe that the best and good states occupy a negligible amount of the search space. On the other hand, over 90\% of the states are bad states. This suggests that the good states occupy small ``pockets'' of the search space. The cost gradient effectively leads SA to identify such pockets and the probabilistic transitions prevent it from getting stuck in local minima. 


\vspace{0.25em}
\noindent \textit{Limitations}. Since we used randomized algorithms, our results
might be sub-optimal. While SA provides statistical guarantees
on the confidence and running time to reach an optimal state, in practice
such guarantees might require multiple runs of algorithms over a long period of time. Additionally, we observe a discrepancy between the validation and test fairness improvements where the test set improves less than the validation set in both \neuronrepair~algorithms. For our experiments, we use Equalized Odds Difference (EOD) only. However, the cost function can be modified accordingly with any fairness metric.

\vspace{0.25em}
\noindent \textit{Threat to Validity}. To address the internal validity and ensure
our finding does not lead to an invalid conclusion, we follow the established
SE guidelines and repeat the experiments 10 times, reporting both the average
and 95\% confidence intervals. In addition, we study the progress of algorithms
during their runs, not just the outcomes. 
To ensure that our results are generalizable and address external validity,
we perform our experiments on seven DNN benchmarks with various architectures.
While similar architectures have been used in the prior works~\cite{10.1109/ICSE48619.2023.00136,zhang2020white,9793943,10.1145/3460319.3464820},
it is an open problem whether these datasets and
DNN models are sufficiently representative for showing the effectiveness of \toolname.

\section{Related Work}
\label{sec:related}
Since the main focus of this paper is unfairness mitigation,
we primarily focus on the prior work on mitigation.

\vspace{0.25 em}
\noindent \textit{A) Pre-processing Techniques} exploit the space of input data
to mitigate fairness defects~\cite{galhotra2022causal,10.1145/2783258.2783311,kamiran2012data,10.5555/3042817.3042973}. Reweighting \cite{kamiran2012data} augments the dataset where the data points in each group-label combination are weighted differently to ensure fairness. The reweighting process minimizes the discrepancy between the observed and expected probabilities of the favorable outcome occurring with the sensitive attribute. \textsc{FairMask}~\cite{10.1109/TSE.2022.3220713} uses
the fact that protected attributes can be inferred by some combinations of other non-protected attributes
(the ``proxy'' problem) to reduce the influence of protected attributes in the inference stage.
\textsc{Fair-SMOTE}~\cite{Chakraborty-FSE'21} used
under-sampling and over-sampling techniques~\cite{10.5555/1622407.1622416} to balance data based on class and sensitive attributes. MAAT \cite{10.1145/3540250.3549093} proposes an ensemble training approach to mitigating unfairness. They propose a framework to train fairness and performance ML models and combine their outputs to make predictions. The performance model is identical to traditional ML algorithms and remains unchanged. For the fairness model, they design a debugging strategy based on prior work \cite{NEURIPS2019_373e4c5d,10.1145/3468264.3468537} that attribute bias in training data to selection bias and label bias. 
LTDD \cite{10.1145/3510003.3510091} hypothesizes that biased features in training data contribute to unfairness. They develop a linear regression-based algorithm that measures the association between non-sensitive and sensitive features in the dataset to obtain unbiased features. The non-sensitive features are then modified to remove the biased parts to become independent of the sensitive features. The sensitive features are dropped from the training dataset after debiasing. 

The works above tamper with the precious training data samples that might lead to unrealistic representations of different protected groups. Additionally, it may not always be feasible to modify training data. Our work is tangential, focusing on post-processing mitigation that does not modify or access the training dataset and works on pre-trained unfair DNNs.

\vspace{0.25 em}
\noindent \textit{B) In-processing techniques} improve fairness during training.  \emph{Adversarial debiasing}~\cite{zhang2018mitigating}, employs adversarial learning to develop a classifier that hinders the ability of adversaries to determine sensitive attributes from ML predictions. Another approach, the \emph{prejudice remover}~\cite{6413831}, incorporates a fairness-centric adjustment into the loss function to balance accuracy and fairness. These strategies necessitate alterations in either the loss function or the model parameters. 
Seldonian~\cite{doi:10.1126/science.aag3311,metevier2019offline,Seldonian} presented a technique that allows users to directly specify (arbitrary) undesirable behaviors as constraints and enforce them during training. 
\textsc{Fairway}~\cite{chakraborty2020fairway} combines pre-processing
and in-processing mitigation techniques to improve fairness. 
\textsc{Parfait-ML}~\cite{tizpaz2022fairness} is a gray-box evolutionary search algorithm that explores the ML hyperparameters to find configurations that minimize
fairness while maintaining an acceptable accuracy. 
While mitigating unfairness during training might be less intrusive than changing the training datasets, it often requires changing the training algorithms. This might only be effective for a particular algorithm and fairness definition. Additionally, the training process is often expensive and might not be feasible for existing systems.

\vspace{0.25 em}
\noindent \textit{C) Post-processing techniques} aim to modify the prediction outcomes of ML models to reduce discrimination~\cite{hardt2016equality,6413831,10.5555/3295222.3295319,tao2022ruler}. Equalized Odds Processing (EOP) \cite{hardt2016equality} modifies the output of a biased binary classifier to improve fairness. EOP solves an optimization problem using a linear program that utilizes the protected attribute, predictions, and true labels from a dataset to create an unbiased predictor. \neuronrepair, however,  does not require training auxiliary models to mitigate unfairness. Additionally, \neuronrepair does not rely on randomization for fairness. Predictions during inference time are deterministic after the neurons are dropped from the DNN. \citet{woodworth2017learning} also show that EOP requires the biased classifier to be Bayes optimal, which is practically impossible to learn from finite samples of data. Zhang and Sun~\cite{10.1145/3540250.3549103} adaptively intervened in the input data features and DNN internal neurons to improve fairness.
\textsc{Care}~\cite{sun2022causality} presented a Particle Swarm Optimization (PSO) algorithm to repair DNNs.  
\textsc{Faire}~\cite{10.1145/3617168} proposes a method to repair unfair DNNs by altering the activations of neurons.
Inspired by program repair techniques, \textsc{Faire} first groups neurons representing protected and non-protected features in the dataset into two categories. To create the two categories, \textsc{Faire} creates a clone of the unfair model and retrains it to predict the protected attribute. By comparing the activations of the corresponding neurons in both networks, \textsc{Faire} determines which neurons represent the protected and non-protected features. Finally, the protected neuron activations are penalized, and the non-protected neuron activations are promoted. \neuronrepair, however, does not retrain the model and does not separately use the protected attribute to identify unfair neurons. We find an unfair subset of neurons given the entire feature vector, thereby accounting for the causal influence of the protected features on the non-protected features in the training dataset. Our analysis considers a group of neurons together rather than identifying the contributions of individual neurons. \textsc{FairNeuron}~\cite{10.1145/3510003.3510087}  proposes an algorithm to mitigate unfairness by selectively retraining neurons in a DNN. On a high level, the algorithm identifies input-output neuron paths corresponding to each sample in the training dataset that induces bias in the prediction. Then, the dataset is split into biased and unbiased subsets corresponding to the neuron activations in the paths. Finally, the algorithm retrains the model with both training data splits by enabling random dropouts in the biased paths and disabling any dropouts in the unbiased paths. While \textsc{FairNeuron} uses dropout to improve fairness, it fundamentally differs from \neuronrepair. We do not retrain the DNN by using the training dataset. Also, \neuronrepair utilizes deterministic dropout during inference time to identify a desirable subset of neurons to drop that has a minimal impact on F1 score but the maximal impact on fairness, rather than dropping neurons randomly.



\section{Conclusion}
\label{sec:conclusion}
In this paper, we tackle the problem of mitigating unfairness in pre-trained DNNs using the dropout method. We showed that the neural dropout problem over the DNN models
is computationally hard and presented \neuronrepair, a family of randomized algorithms to efficiently
and effectively improve the fairness of DNNs.
Our experiments showed that \toolname can identify an ideal subset of neurons to drop that disparately contribute to unfairness (leading to up to 69\% fairness improvement) and outperform a state-of-the-art post-processing bias mitigator. 

For future work, there are a few exciting directions. First, we can analyze the top $k$ states found during the run instead of choosing the best state.
Such a strategy could reduce the difference in improvement observed between the validation and test sets.
We can extend to multi-valued protected attributes (e.g., age groups) or optimize for more than one protected attribute at a time (e.g., race and sex). 
Second, we plan to leverage dropouts in the fine-tuning process to understand the effect of randomized algorithms on mitigating the unfairness and toxicity of large language models. Lastly, alternate strategies like genetic algorithms may be used to solve the combinatorial optimization problem of neuron dropout. We use randomized algorithms for their simplicity in efficiently solving combinatorial optimization problems. The primary difference between SA/RW and genetic algorithms is that genetic algorithms create a population of candidate solutions at each step instead of a single solution. This approach can be promising as it enables us to explore more states of the search space simultaneously.

\section*{Data Availability}
Our open-source tool \neuronrepair~with all experimental subjects are
publicly accessible on Zenodo \cite{dasu_2024_12662049} and \href{https://github.com/vdasu/neufair}{GitHub}.

\begin{acks}
This material is based upon work supported by the National
Science Foundation under Grant No. CNS-2230060 and CNS-2230061.
\end{acks}

\bibliographystyle{ACM-Reference-Format}
\bibliography{main}

\end{document}